\documentclass[letterpaper, 10 pt, conference]{ieeeconf}  

\IEEEoverridecommandlockouts                              

\usepackage{amsmath} 
\usepackage{amssymb}  

\usepackage{graphicx}

\usepackage{xspace}
\makeatletter
\DeclareRobustCommand\onedot{\futurelet\@let@token\@onedot}
\def\@onedot{\ifx\@let@token.\else.\null\fi\xspace}
\def\eg{\emph{e.g}\onedot} 
\def\ie{\emph{i.e}\onedot}

\def\etal{\emph{et al}\onedot}

\makeatother

\title{\LARGE \bf
A Unified Formulation for Visual Odometry*
}

\author{Georges Younes$^{1}$, Daniel Asmar$^{2}$, and John Zelek$^{1}$
\thanks{*This work was funded by the University Research Board (UBR) at the American University of Beirut, and the Canadian National Science Research Council (NSERC).}
\thanks{$^{1}$Georges Younes and John Zelek are with the Systems Design Engineering Department, University of Waterloo, N2L 3G1 Waterloo, Canada.
        {\tt\small gyounes@uwaterloo.ca, jzelek@uwaterloo.ca}.}%
\thanks{$^{2}$Daniel Asmar is with the Department of Mechanical Engineering, American University of Beirut, 1107 2020, Beirut, Lebanon
        {\tt\small da20@aub.edu.lb}}.%
}

\begin{document}

\maketitle
\thispagestyle{empty}
\pagestyle{empty}

\begin{abstract}
Monocular Odometry systems can be broadly categorized as being either \textit{Direct}, \textit{Indirect}, or a hybrid of both. While Indirect systems process an alternative image representation to compute geometric residuals, Direct methods process the image pixels \textit{directly} to generate photometric residuals. Both paradigms have distinct but often complementary properties.  This paper presents a Unified Formulation for Visual Odometry, referred to as UFVO, with the following  key contributions: (1) a tight coupling of photometric (Direct) and geometric (Indirect) measurements using a joint multi-objective optimization, (2) the use of a utility function as a decision maker that incorporates prior knowledge on both paradigms, (3) descriptor sharing, where a feature can have more than one type of descriptor and its different descriptors are used for tracking and mapping, (4) the depth estimation of both corner features and pixel features within the same map using an inverse depth parametrization, and (5) a corner and pixel selection strategy that extracts both types of information, while promoting a uniform distribution over the image domain. Experiments show that our proposed system can handle large inter-frame motions, inherits the sub-pixel accuracy of direct methods, can run efficiently in real-time, can generate an Indirect map representation at a marginal computational cost when compared to traditional Indirect systems, all while outperforming state of the art in Direct, Indirect and hybrid systems. 
\end{abstract}

\section{Introduction}
Monocular Odometry is the process of analyzing information extracted from the images of a camera to recover the camera's motion within a local map of its surroundings. 

A key component in the design of a Visual Odometry (VO) system is the information extracted from the images. In Indirect systems, the images are pre-processed to extract salient features with their associated descriptors, which are then used as input to VO. In contrast, Direct systems use the brightness value of a pixel and its surrounding pixels as inputs to VO. Direct and Indirect methods exhibit complementary behavior, where the shortcomings of each are mitigated in the other.
For example, Direct methods have a natural resilience to texture-less environments since they can sample any pixel in the image, whereas Indirect methods fail in such environments due to corner deprivation. On the other hand, Direct methods suffer from a high degree of non-convexity \cite{engel_2016_ARXIV}, limiting their operation to scenarios where the camera motion is accurately predictable, and subsequently fail when their motion model is violated; 
in contrast, Indirect methods can easily cope with such violations. Another key difference is the amount of precision each system can achieve; Direct methods can integrate over the image domain resulting in a sub-pixel level of precision, whereas Indirect methods operate on a discretized image space limited by the image resolution, thereby suffer from relatively lower precision. While this issue can be reduced with sub-pixel refinement, the refinement process itself is based on interpolating the feature location using its pixel representation (\ie , Direct alignment). For further discussion and analysis on the differences between both paradigms, the interested reader is referred to \cite{younes_2016_ARXiv}, \cite{younes_2018_arxiv} and \cite{yang_2018_RAL}.
\begin{figure}[!tb]
	\centering
	\includegraphics[trim={0.0cm 0cm 0cm 0.0cm},clip,width=0.48\textwidth]{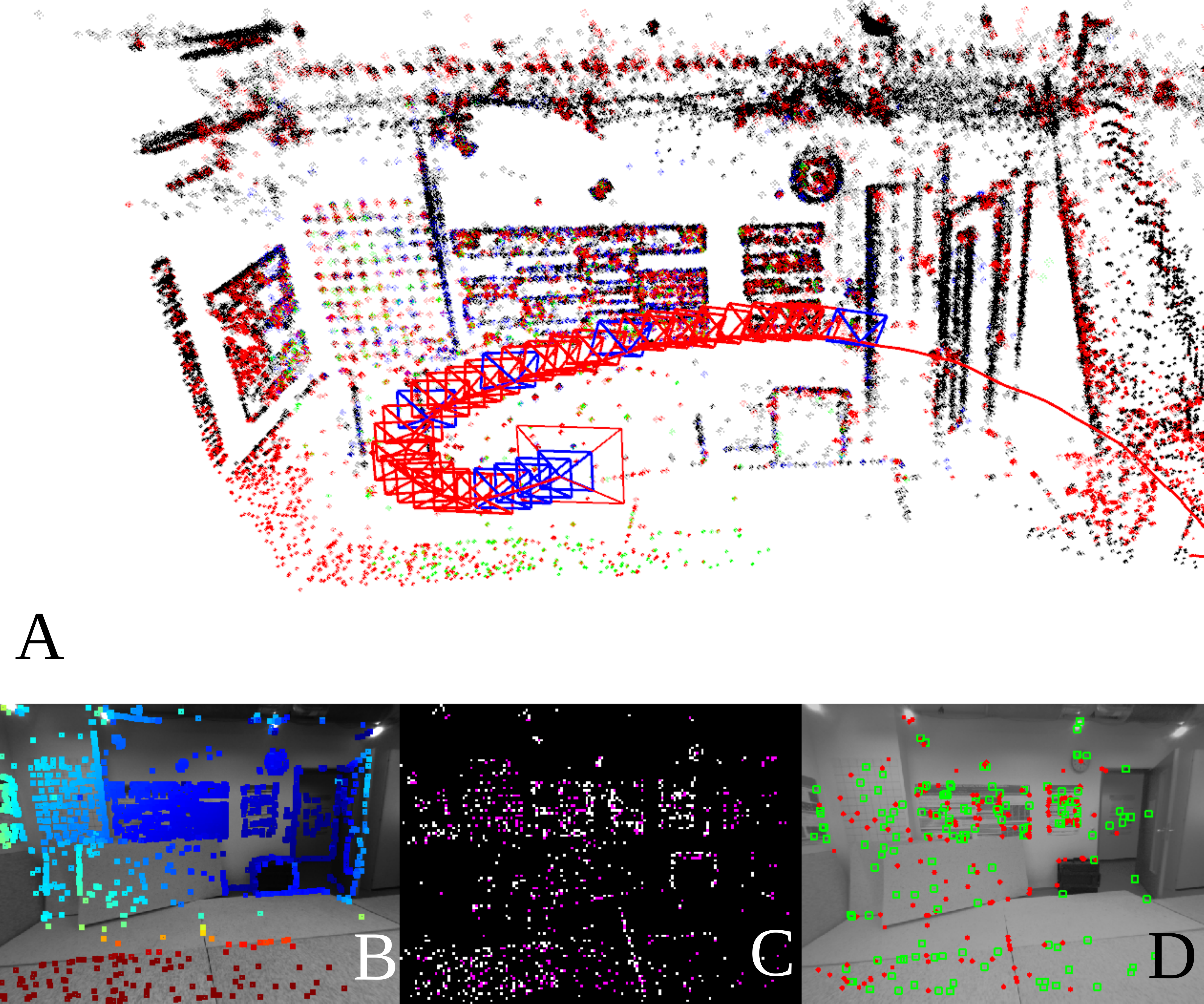}
	\caption{ \textbf{UFVO}: A$-$ shows the 3D recovered map and the
	different types of points used: green points contribute to both
	geometric and photometric residuals, red points geometric
	residuals only, blue points photometric residuals only and black
	points are marginalized (do not contribute any residuals). The blue squares are hybrid keyframes
	(contains both geometric and photometric information), whereas red
	squares are Indirect Keyframes whose photometric data was
	marginalized. B$-$ shows the projected depth map of all active
	points. C$-$ shows the occupancy grid which we use to ensure a
	homogeneously distributed map point sampling process. The white
	squares correspond to the projected map points while the magenta
	squares represent newly extracted features from the newest
	keyframe. Finally D$-$ shows the inlier geometric features (active
	in green and marginalized in red)  used during tracking (Figure
	better seen in color).}
	\label{fig:Main}
\end{figure}
In an effort to capitalize on the advantages of Direct and Indirect systems, we propose a \textit{unified formulation} where both types of information are used to estimate the camera's motion. At the core of our system, a joint multi-objective optimization minimizes both types of residuals at frame-rate. To control the contribution of each type to the optimization, we introduce a utility function that incorporates our prior knowledge on the behaviour of both residuals.
To achieve optimal performance, we use the inverse depth parametrization for both types of information, which allows us to estimate the depth of both types concurrently. A corner and pixel sampling strategy is also suggested to ensure that both types of residuals are uniformly distributed across the image without redundant information. Fig. \ref{fig:Main} shows an overview of the different proposed components. We call our system a \textit{Unified formulation for Visual Odometry} or UFVO for short. While UFVO is tailored to a monocular system, all of the ideas put forward can be applied to stereo and RGB-D systems.
\section{Background}
\subsection{Nomenclature}
Before getting into further details, 
a nomenclature must be established. 
In particular we refer to the image locations from which measurements are taken as features. This means that we consider corners and pixel locations as features that can be used interchangeably as Direct or Indirect features. We also employ the word \textit{descriptor} for both paradigms, where an Indirect feature descriptor is a high dimensional vector computed from the area surrounding the feature (\textit{e.g.} ORB \cite{rublee_2011_ICCV}), and a Direct feature descriptor is a local patch of pixel intensities surrounding the feature. Finally, we refer to geometric residual as the 2-D geometric distance between an indirect feature location and its associated map point projection on the image plane. In contrast, we refer to photometric residual as the intensity difference between a direct feature descriptor and a patch of pixel intensities at the location where the feature projects to in the image plane.

\subsection{Related Work}
Various systems have previously attempted to leverage the advantages of Direct and Indirect paradigms, resulting in what is referred to as hybrid methods. The first hybrid method was suggested in
\cite{forster_2017_tro}, where corners were extracted at frame rate and data association was performed using a direct image alignment process. While \cite{forster_2017_tro} achieved significant reductions in the required computational cost, their system could not handle large inter-frame motions; as a result, their work was limited to high frame rate cameras, which ensured small frame-to-frame motion.\\
Other hybrid approaches include the works of \cite{krombach_2016_ICIAS} and \cite{jellal_2016_ECMR}; both systems adopted an indirect formulation on a frame-to-frame basis and subsequently applied a Direct formulation on keyframes. However, since they rely on Indirect methods as a front-end, they both suffer in texture-deprived environments; to mitigate this issue,
they resorted to stereo cameras.

More recently, several hybrid approaches were proposed.
In Younes \etal \cite{younes_2018_arxiv}, a Direct formulation is employed as a front-end with an auxiliary Indirect intervention when failure is detected. While this led to significant performance improvements, it required an explicit failure detection mechanism and a separate process to build and maintain an indirect map representation.\\
Gao \etal \cite{Gao_2018_iros} suggested using both corners and pixels as Direct features, where both contribute photometric residuals; however, geometric residuals are not used. The ORB descriptors associated with the corners are only used to detect loop closure. Nevertheless, the introduction of corners as Direct features affects the odometry performance. The reason is that corners have high intensity gradients along two directions, and as such are not susceptible to drift along the edge of an intensity gradient; in contrast, their pixel feature counterparts can drift along edges. While this is expected to improve performance, its results where different across various sequences. 

In Lee and Civera \cite{lee_2019_ral}, the state of the art in both Direct (DSO \cite{engel_2016_ARXIV}) and Indirect (ORB SLAM 2\cite{mur-artal_2015_TRO}) are cascaded, where DSO is used for odometry and its marginalized keyframes are fed to ORB SLAM 2 for further optimization. Aside the inefficiency of running both systems sequentially, the Indirect residuals do not contribute towards real-time odometry unless loop closure is detected, limiting the odometry performance to that of \cite{engel_2016_ARXIV}.

Finally, the closest in spirit to our work is that of Yu \etal \cite{yu_2018_access}, where geometric and photometric residuals are combined in a joint optimization. However, the residuals are treated as completely independent entities, requiring separate processes to extract and maintain each type of residual, irrespective of the other, and the optimization is used as a black box without incorporating crucial priors on the residuals' behavior. 
 

UFVO addresses all of the aforementioned limitations by using both types of residuals at frame-rate. The contribution of our work is a novel monocular odometry system that:
\begin{enumerate}
	\item can operate in texture-deprived environments,
	\item can handle large inter-frame motions (as long as a sufficient number of geometric residuals are established),
	\item does not require a separate process to generate two different types of maps,
	\item incorporates prior information on the various feature types via a utility function in the optimization.   
\end{enumerate}

\section{Notation}
Throughout the remainder of this document, 
we denote matrices by bold upper case letters \textbf{M}, vectors by bold lower case letters \textbf{v}, camera poses as \textbf{$\mathbf{T} \in SE(3)$} with their associated Lie element in the groups' tangent space as {$\mathbf{\hat\xi} \in se(3)$}. A 3D point in the world coordinate frame is denoted as \textbf{x}$\in \Re^3$, its coordinates in the camera frame as $\mathbf{\tilde{x}}=\mathbf{T} \mathbf{x}$, and its projection on the image plane as $\textbf{p}=(u,v)^T$. The projection from 3D to 2D is denoted as $\Pi(\mathbf{c},\mathbf{x}): \Re^3 \rightarrow \Re^2$ and its inverse $\Pi^{-1}(\mathbf{c},\textbf{p},d):\Re^2\rightarrow \Re^3$ where \textbf{c} and d represent the camera intrinsics and the points' depth respectively. $\mathcal{L}(a,b):I(\mathbf{p})\mapsto e^{-a}(I(\mathbf{p})-b) $ is an affine brightness transfer function that models the exposure change in the entire image and $I(\mathbf{p})$ is the pixel intensity value at $\mathbf{p}$. To simplify the representation we define $\mathbf{\mathbf{\xi}}:=(\mathbf{\hat\xi},\mathcal{L})$ as the set of variables over which we perform the camera motion optimization. 
We define the operator $\boxplus:se(3)\times SE(3) \rightarrow SE(3)$ as $\hat\xi\boxplus\mathbf{T}=e^{\hat\xi}\mathbf{T}$.
The incremental updates over $\xi$ are then defined as
$\delta\mathbf{\xi}\oplus\mathbf{\xi}=(log(\delta\mathbf{\hat\xi}\boxplus e^{\mathbf{\hat\xi}}),a+\delta a,b+\delta b)$.
Finally a subscript $p$ is assigned for photometric measurements and $g$ for geometric measurements. 

\section{Proposed System}
\subsection{Feature types}
UFVO concurrently uses both photometric and geometric residuals at frame-rate. For this purpose, we make the distinction between two types of features: salient corner features, and pixel features. 

\subsubsection{Corner features} are FAST \cite{rosten_2006_ECCV} corners extracted at $\mathbf{p}$, associated with a Shi-Tomasi score \cite{shi_1994_CVPR} that reflects their saliency as a corner, an ORB descriptor \cite{rublee_2011_ICCV}, and a patch of pixels surrounding the point \textbf{p}. Corner features contribute two types of residuals during motion estimation: a geometric residual using its associated descriptor, and a photometric residual using its pixels patch. 

\subsubsection{Pixel features} are sampled from the images at any location \textbf{p} that is not a corner and has sufficient intensity gradient; they are only associated with a patch of pixels; therefore, pixel features only contribute photometric residuals during motion estimation.

\begin{figure}[!tb]
	\centering
	\includegraphics[trim={0.0cm 0cm 0 0.0cm},clip,width=0.4\textwidth]{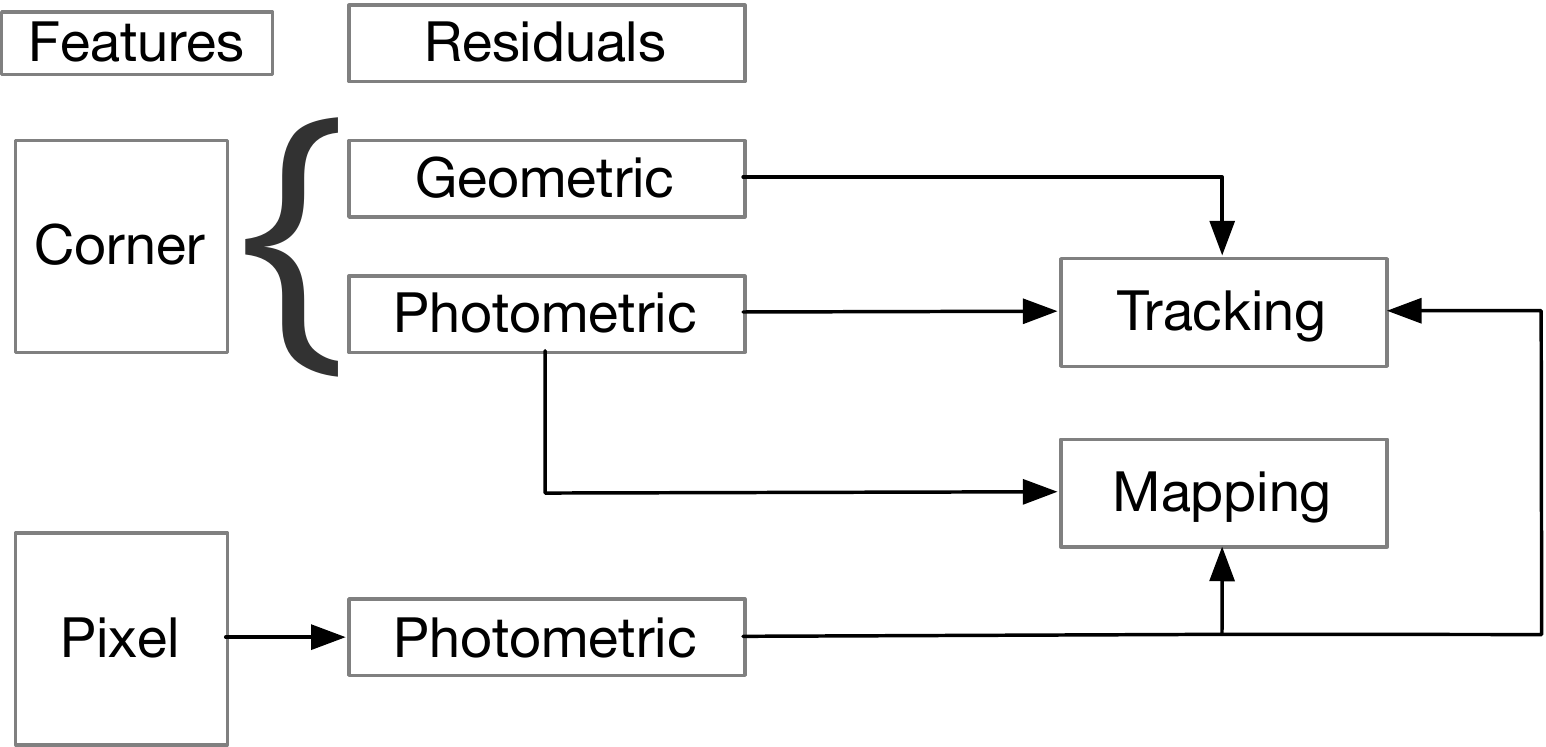}
	\caption{Summary of the feature types, their associated residuals, and their usage in UFVO.}
	\label{fig:Cont}
\end{figure}

The types of residuals each feature type contributes in tracking and mapping are summarized in Fig. \ref{fig:Cont}.
Features are classified as either:
\begin{enumerate}
	\item \textit{Candidate}: new features whose depth estimates have not converged; they contribute to neither tracking nor mapping.
	\item Active: features with depth estimates that contribute to both tracking and mapping.
	\item Marginalized: features that went outside the current field of view or features that belong to marginalized keyframes.
	\item Outliers: features with high residuals or corners that frequently failed to match other features. 
\end{enumerate}

\subsection{System Overview}
\begin{figure*}[!ht]
	\centering
	\includegraphics[trim={0.0cm 0cm 0 0.0cm},clip,width=0.9\textwidth]{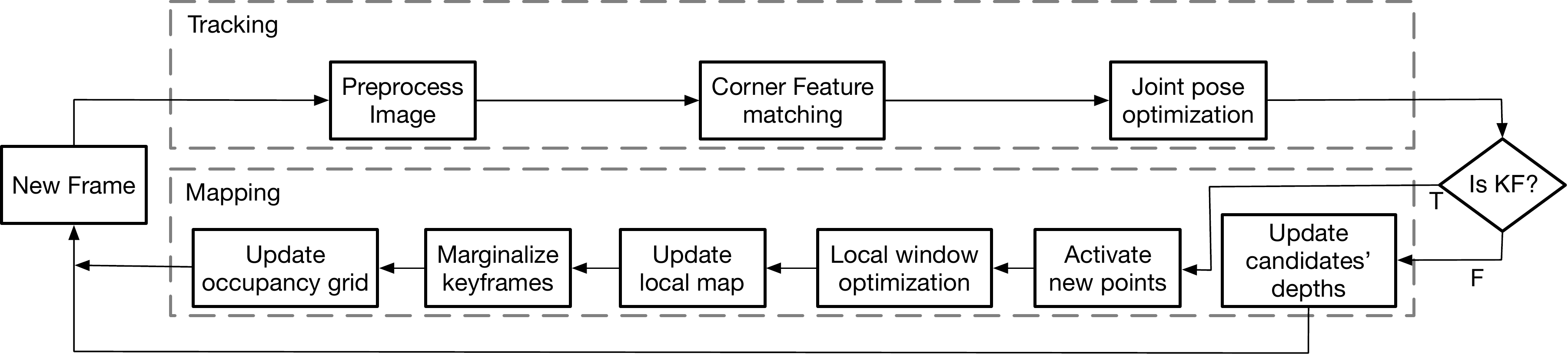}
	\caption{Overview of UFVO's tracking and mapping threads. A new frame is first pre-processed to compute pyramid levels and extract corner features. Corner features matching then takes place before a joint pose optimization. UFVO then updates an occupancy grid over the last added keyframe which records the locations of active corners and pixel features. The frame is then passed to the mapping thread and a decision is made whether it should be a keyframe or not. If it is not selected as a keyframe, it is used to update the depth estimates of the local map's candidate points. Otherwise, candidate points from the local map are activated and a local photometric optimization takes place; the local map is then updated with the optimized variables and old keyframes with their associated points are marginalized.}
	\label{fig:Overview}
\end{figure*}
UFVO is an odometry system that operates on two threads: tracking and local mapping. Fig. \ref{fig:Overview} depicts an overview of the system's operation, which starts by processing new frames to create a pyramidal image representation, from which both corners features are first sampled. A constant velocity motion model is then used to define a search window for corner feature matching, which are used to compute the geometric residuals. A joint multi-objective pyramidal image alignment then minimizes both geometric and photometric residuals associated with the two types of features over the new frame's pose. The frame is then passed to the mapping thread where a keyframe selection criterion (as described in \cite{engel_2016_ARXIV}) is employed. If the frame was not flagged as a keyframe, all the candidate points' depth estimates are updated using their photometric residuals in an inverse depth formulation and the system awaits a new frame.

On the other hand, if the frame was deemed a keyframe, a 2D occupancy grid is first generated over the image by projecting the local map points to the new keyframe; each map point occupies a $3x3$ pixels area in the grid. A subset of the candidate features from the previous keyframes are then activated such that the new map points project at empty grid locations. New \textit{Candidate} corner features are then sampled at the remaining empty grids before a local photometric bundle adjustment takes place which minimizes the photometric residuals of all features in the active window. Note that we do not include the geometric residuals in this optimization; their use during tracking ensures that the state estimates are as close as possible to their global minima; hence, there is no added value of including them. Furthermore, since the geometric observation models' precision is limited, including them would actually cause jitter around the minimum.
 
Outlier Direct and Indirect features are then removed from the system. The local map is then updated with the new measurements and a computationally cheap structure only optimization is applied to the marginalized Indirect features that are part of the local map.
Finally, old keyframes are marginalized from the local map. 

\subsection {Corner and pixel features sampling and activation}
\subsubsection{Feature sampling}
when a new frame is acquired, a pyramidal representation is created over which we apply a pyramidal image alignment; however, unlike
\cite{mur-artal_2015_TRO} or most Indirect methods in the literature, we only extract
Indirect features at the highest image resolution. Since Indirect features in UFVO are only tracked in a local set of keyframes, which are relatively close to each other (\ie , don't exhibit significant variations in their scale), extracting features from one pyramid level allows us to save on the computational cost typically associated with extracting Indirect features without significantly compromising performance.

Since corners contribute to both types of residuals, we avoid sampling pixel features at corner locations; therefore, we sample pixel features at \textit{non-corner} locations with sufficient intensity gradient.

\subsubsection{Feature activation}
when a keyframe is marginalized, a subset of the candidate features from the previous keyframes are activated to take over in its place. Our feature activation policy is designed to enforce the following priorities in order:
\begin{enumerate}
	\item To minimize redundant information, features should not overlap with other types of features. 
	\item Ensure maximum saliency for the new Indirect features. 
	\item To maintain constant computational cost, add a fixed number of new Direct and Indirect Candidates.
\end{enumerate}

To enforce the activation policy and ensure a homogeneous feature
distribution, we employ a coarse 2D occupancy grid over the latest keyframe, such that the grid is populated with the projection of the current map on the keyframe with each point occupying a $3x3$ area in the occupancy grid. 

Since corners are scarce, we prioritize their activation, that is, we employ a two stage activation process: the first sorts corner features in a descending order according to their Shi-Tomasi score, and activates a fixed number of the strongest corners from unoccupied grid cells. The second stage activates pixel features at locations different than the newly activated corners, and that maximizes the distance to any other feature in the keyframe.

\begin{figure}[ht]
	\centering
	\includegraphics[trim={0.0cm 0cm 0 0.0cm},clip,width=0.5\textwidth]{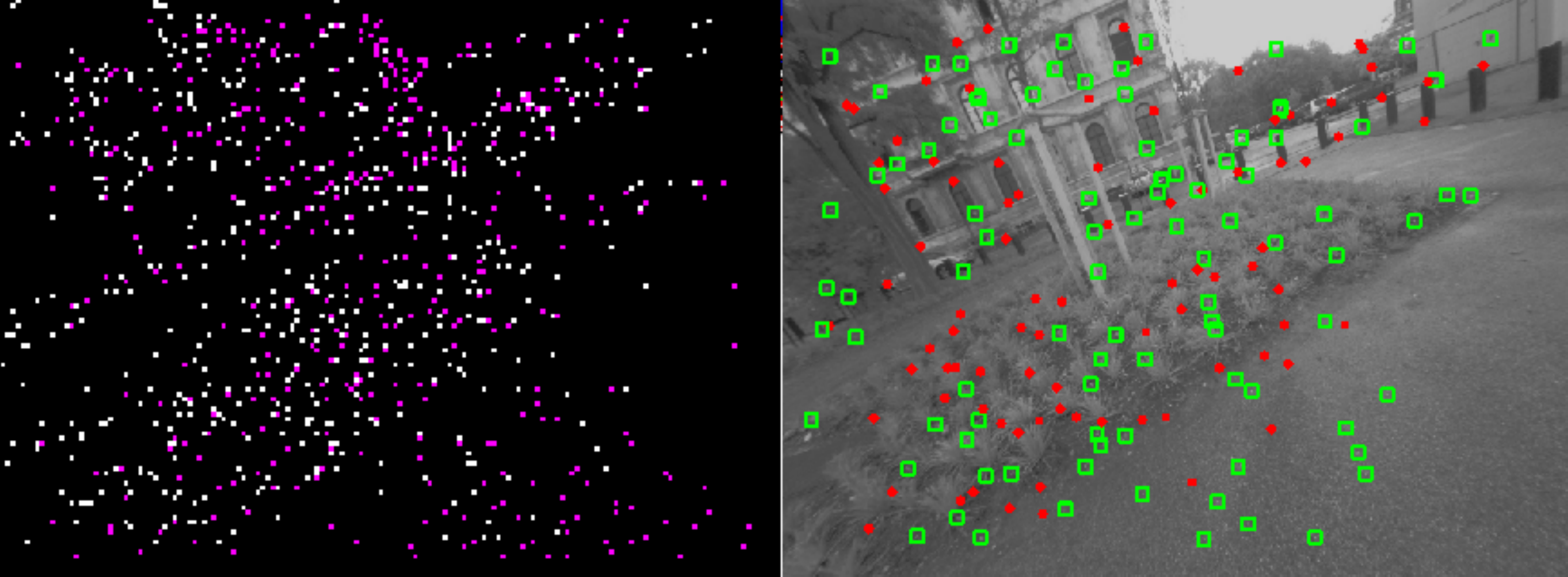}
	\caption{Occupancy grid (left), and its associated keyframe's image (right). The white squares in the occupancy grid represent current map points and the magenta squares represent newly added ones. As for the right image, the green squares represent indirect active map point matches, and the red dots represent marginalized indirect feature matches. 
	}
	\label{fig:Occup}
\end{figure}

Fig. \ref{fig:Occup} demonstrates the effectiveness of our sampling and activation strategy by showing an example of the occupancy grid (left) alongside its keyframe's image (right). The occupancy grid (left) shows the current map points in white and the newly added map points in magenta. It can be seen that there is no overlap between old and new points, the points are homogeneously distributed throughout the frame.

\subsection {Joint Multi-Objective Image alignment}
The core component of UFVO is a joint optimization that minimizes an energy functional, which combines both types of residuals, over the relative transformation $\xi$ relating the current frame to last added keyframe. The joint optimization is best described as:
\begin{equation}
\label{eq:EnergyFunc}
\underset{\xi}{\operatorname{argmin}}(\mathbf{e}_p(\xi),\mathbf{e}_g(\xi)).
\end{equation}
The optimization process itself must be computationally efficient, delivers a single pareto optimal solution, and capable of achieving superior performance than both individual frameworks. 
While the research community has provided an ample amount of methods for Multi-Objective optimizations, few meet the harsh constraints of real-time performance, and allow for explicit a priori articulation of preferences. One optimization method that meets the aforementioned requirements is the \textit{Weighted Sum Scalarization} method \cite{Marler_2010_SMO}, that transforms the optimization of \eqref{eq:EnergyFunc} to:
\begin{equation}
 \label{eq:Scalarized}
 \underset{\xi}{\operatorname{argmin}}\sum(\alpha_1\mathbf{e}_p(\xi),\alpha_2\mathbf{e}_g(\xi)),
\end{equation}
where $\alpha_1$ and $\alpha_2$ represent the contribution of each
residual type to the final solution. For simplicity, we reformulate the problem using $K=\frac{\alpha_2}{\alpha_1}$, which represents the weight of the geometric residuals relative to the photometric residuals; \textit{e.g.} $K=2$ assigns twice as much importance to the geometric residuals than to the photometric residuals.

For this weighing scheme to have any sense, both energies must be dimensionless, and normalized such that imbalances in the numbers of the two residuals does not inherently bias the solution. The Huber norm is also used to account for outliers. The joint energy functional becomes:
\begin{equation}
\label{eq:ScalarizedRed}
\underset{\xi}{\operatorname{argmin}}\:\mathbf{e}(\xi)=\underset{\xi}{\operatorname{argmin}}\left[\frac{\|\mathbf{e}_p(\xi)\|_\gamma}{n_p\sigma_p^2}+K\frac{\|\mathbf{e}_g(\xi)\|_\gamma}{n_g\sigma_g^2}\right],
\end{equation}
where $n$ is the count of each feature type, $\sigma^2$ is the residual's variance, $\parallel\cdot\parallel_\gamma$ is the Huber norm, and the energy per feature type is defined as:
\begin{equation}
\label{eq:errorval}
 \mathbf{e}(\xi)=\mathbf{r}^TW\mathbf{r},
\end{equation}
$\mathbf{r}$ is the vector of stacked residuals per feature type and $W$ is a weight matrix that will be further discussed by the end of this section.

We seek an optimal solution ${\bar\xi}$ that minimizes
\eqref{eq:ScalarizedRed}; however, $\mathbf{r}$ is non-linear, therefore we linearize it with a first order Taylor
expansion around the initial estimate $\xi$, with a perturbation
$\delta\xi$, that is:
\begin{equation}
\label{eq:linearize}
\mathbf{r(\xi\oplus\delta\xi)}\simeq \mathbf{r}(\xi)+\mathbf{J}\delta\xi,
\end{equation}
where $\mathbf{J}=\frac{\partial\mathbf{r}}{\partial\xi}$.
If we replace $\mathbf{r}$ in \eqref{eq:errorval} by its linearized value from \eqref{eq:linearize}, substitute the result in
\eqref{eq:ScalarizedRed}, differentiate the result with respect to $\xi$, and set it equal to zero, we arrive at the step increment equation $\delta\xi \in \Re^8$ of the joint optimization:
\begin{multline}
\label{eq:stepupdate}
\delta\xi=-\left[  \left(\frac{\mathbf{J}^t\mathbf{W}\mathbf{J}}{n\sigma^2}\right)_p + \left(  \frac{K\mathbf{J}^t\mathbf{W}\mathbf{J}}{n\sigma^2}     \right)_g\right]^{-1} \\ \left[\left(\frac{\mathbf{J}^t\mathbf{W}\mathbf{r}}{n\sigma^2}    \right)_p+\left(\frac{K\mathbf{J}^t\mathbf{W}\mathbf{r}}{n\sigma^2}\right)_g \right],
\end{multline}
which we iteratively apply using $\xi=\delta\xi\oplus\xi$ in a
Levenberg-Marquardt formulation, until convergence. The optimization is repeated from the coarsest to the finest pyramid levels, where the result of each level is used as an initialization to the subsequent one. At the end of each pyramid level, we remove outliers and update the variable $K$ according to \ref{sec:utility}. 

The photometric residual $r_p\in \Re$ per feature can be found by evaluating:
\begin{equation}
r_p= \sum_{p\in\mathcal{N}_p}\left[(I_j[\mathbf{p'}]-b_j) -\frac{t_je^{a_j}}{t_ie^{a_i}}(I_i[\mathbf{p}]-b_i)\right],
\end{equation}
where $\mathcal{N}_p$ is the neighboring pixels of the feature at $\mathbf{p}$,
the subscript $i$ denote the reference keyframe, and $j$ the current
frame; $t$ is the image exposure time which is set to 1 if not-available,
and $\mathbf{p'}$ is the projection of a feature from the reference
keyframe to the current frame, which is found using: 
\begin{equation}
\mathbf{p'}=\Pi(\mathbf{c},e^{\hat\xi}\Pi^{-1}(\mathbf{c},\mathbf{p},d))
\end{equation} 
where $\hat\xi$ is the relative transformation from the reference keyframe to the new frame. Note that we compute the photometric residual for both types of features (corners and pixels).
The geometric residual $r_g\in \Re^2$ per corner feature is defined as:
\begin{equation}
\mathbf{r}_g=\mathbf{p}' -\mathbf{obs},
\end{equation}
where $\mathbf{obs}\in \Re^2 $ is the corners' matched location in the new image, found through descriptor matching. 
\label{sec:resids}
Regarding the weight $W$ matrices, the photometric weight is defined as $w_p=h_w(\gamma_p)$ where:
\begin{equation}
\label{eq:huberweight}
h_w=\left\{ \begin{array}{cc} 1 & if\, e < \gamma^2 \\ \frac{\gamma}{\sqrt{e}} & if \; e\geq\gamma^2 \end{array} \right. ,
\end{equation} is the Huber weight function. 
As for the geometric weight, we combine two weighing factors: a Huber
weight as defined in \eqref{eq:huberweight}, and a variance weight associated with the variance of the corners' depth estimate:
\begin{equation}
w_d=\frac{\frac{1}{\sigma^2_d}}{max\left(\frac{1}{\sigma_d^2}\right) },
\end{equation} with $max(\frac{1}{\sigma_d^2})$ the maximum $\frac{1}{\sigma_d^2}$ in the current frame, which down-weighs features with high depth variance. The final geometric weight is then found as $W_g=w_dh_w(\gamma_g)$.

The Jacobians required to solve \eqref{eq:stepupdate} are shown in the appendix.

\subsection{Utility function}
\label{sec:utility}
Due to the high non-convexity of the Direct formulation, erroneous or initialization points far from the optimum, cause the Direct optimization to converge to a local minimum, far from the actual solution. While Indirect methods are robust to such initializing points, they tend to flatten around the actual solution due to their discretization of the image space. The interested reader is referred to \cite{younes_2018_arxiv} for an experiment on the matter. Ideally, an optimization process would follow the descent direction of the Indirect formulation until it reaches a pose estimate within the local concavity of the actual solution, after which it would follow the descent direction along the Direct formulation.

The introduction of $K$ in \eqref{eq:ScalarizedRed} allows us to express such a-priori preference within the optimization. As $k\rightarrow0$ the optimization discards geometric residuals, whereas as $k\gg$, geometric residuals dominate. Therefore we seek a function that controls $K$ such that the descent direction behaves as described earlier.  Furthermore, the geometric residuals tend to be unreliable in texture-deprived environments, therefore $K$ should be $\propto$ number of matches. We heuristically design the following logistic utility function:
\begin{equation}
\label{eq:Utility}
K=\frac{5e^{-2l}}{1+e^{\frac{30-N_g}{4}}},
\end{equation} 
where $l$ is the pyramid level at which the optimization is taking place, and $N_g$ is the number of current inlier geometric matches. While the number of iterations does not explicitly appear in \ref{eq:Utility}, it is embedded within $l$; as the optimization progresses sequentially from a pyramid level to another, the optimization follows the descent direction of the geometric residuals,with a decay induced by \eqref{eq:Utility} that down-weighs the contribution of the geometric residuals as the solution approaches its final state. $K$ also penalizes the Indirect energy functional at low number of matches, allowing UFVO to naturally handle texture-deprived environments.

\subsection{Mapping}
\subsubsection{Map representation}
Unlike typical Indirect formulations found in the literature \cite{mur-artal_2015_TRO},\cite{younes_2018_arxiv},\cite{klein_2007_ISMAR}, etc., we adopt for our Indirect features an inverse depth parametrization, which allows us to circumvent the need for a separate multi-view geometric triangulation process that is notorious for its numerical instability for observations separated by small baselines. Instead, we exploit the Direct pixel descriptors associated with our Indirect corner features. Aside its numerical stability, this is also advantageous in terms of computational resources, as it allows us to compute the depth of Indirect features at virtually no extra computational cost.

\subsubsection{Local Map}
Our local map is made of a moving window of keyframes in which we distinguish between two types of keyframes: 
\begin{itemize}
	\item hybrid keyframes: a fixed-size set of keyframes that contains Direct and Indirect residuals, over which a photometric bundle adjustment optimizes both types of features using their photometric measurements. 
	\item Indirect keyframes: previously hybrid keyframes whose photometric data was marginalized, but still share Indirect features with the latest frame. 
\end{itemize} 
As new keyframes are added, hybrid keyframes are removed by marginalization using the Schur complement. On the other hand, Indirect keyframes are dropped from the local map once they no longer share Indirect features with the current camera frame.

To maintain the integrity of marginalized Indirect points we resort to a structure-only optimization that refines their depth estimates with new observations; however, one should note that the use of marginalized indirect features is restricted to features that are still in the local map. Furthermore, their use, and the structure only optimization for that matter, is optional; however, we found that using them increases the system's performance as it allows previously marginalized reliable data to influence the current state of the system, thereby reducing drift.



\section{Evaluation}
Our evaluation includes a computational cost analysis and a thorough evaluation of our system on various sequences from the TUM Mono dataset \cite{engel2_2016_arxiv}, covering a wide range of full camera exploration scenarios with fast motions, texture-deprived environments, short and long paths in indoor and outdoor scenes.

Since our system is a monocular system, we also report on the state-of-the-art in monocular Direct (DSO \cite{engel_2016_ARXIV}), Indirect (ORB SLAM2 \cite{mur-artal_2015_TRO}) and Hybrid (LDSO \cite{Gao_2018_iros} and LCSD \cite{lee_2019_ral}). For fairness of comparison and similar to \cite{engel_2016_ARXIV} and \cite{younes_2018_arxiv}, we evaluate ORB SLAM2, LDSO and LCSD as odometry systems by disabling their global loop closure modules.
We don't compare against SVO 2 \cite{forster_2017_tro} since it failed on most sequences, and when it didn't fail it returned errors of an order of magnitude larger than the other systems; for reported results on \cite{forster_2017_tro} the reader is referred to \cite{yang_2018_RAL}.  Finally we don't compare against our own previous hybrid system (FDMO \cite{younes_2018_arxiv}) since UFVO is a much more efficient extension of FDMO that runs on every frame, and it naturally outperforms FDMO.

\subsection{Computational Cost}
We analyze the computational cost associated with the major components of UFVO on an Intel Core i7-8700K CPU @ 3.70GHz CPU; no GPU acceleration was used. The time required by each process is summarized in Tab. \ref{tab:CompTime}.
\begin{table}[!hbt]
	\renewcommand{\arraystretch}{1.3}
	\caption{Computational cost of different components of UFVO in ms.
		\label{tab:CompTime}}
	\centering
	\begin{tabular}{l c }
		\hline
		Process & mean time (ms)\\
		\hline
		Direct data preparation and Image Pyramids & 3.81 \\
		\hline
		Features and Descriptors Extraction & 3.38 \\
		\hline
		Feature Matching & 2.52 \\
		\hline
		Joint Optimization &4.51  \\
		\hline
		Occupancy map Update& 2.62\\
		\hline
		Candidate Points Depth Update & 3.5\\
		\hline
		New map point initialization & 4.27\\
		\hline
		Photometric BA & 50.6\\
		\hline
		Local Map Update & 4.62\\
		\hline
		Structure only optimization (optional) & 7.32\\
		
		\hline
	\end{tabular}
\end{table}
In total, UFVO requires on average 14.2 ms per frame for tracking and 72.9 ms on average per keyframe for mapping both Direct and Indirect map representations. In contrast, DSO requires 4.46 ms for tracking and 46.77 for mapping a Direct representation while ORB SLAM 2 requires 19.08 for tracking and 104.44 ms for mapping an Indirect map only.  

\subsection{Quantitative Evaluation}
To evaluate the performance of our system we report on the alignment error $e_{align}$, as described in \cite{engel2_2016_arxiv}, which encompasses the effects of translation, rotation and scale drifts accumulated over an entire path. Each sequence is repeated ten times, with the alignment error measured for each sequence. We report on the alignment error of UFVO (ours), LCSD \cite{lee_2019_ral}, LDSO \cite{Gao_2018_iros}, DSO \cite{engel_2016_ARXIV} and ORB SLAM 2 \cite{mur-artal_2015_TRO}.

\begin{figure*}[!htb]
	\centering
    \includegraphics[trim={0.3cm 0cm 0.3cm 0.0cm},clip,width=\textwidth]{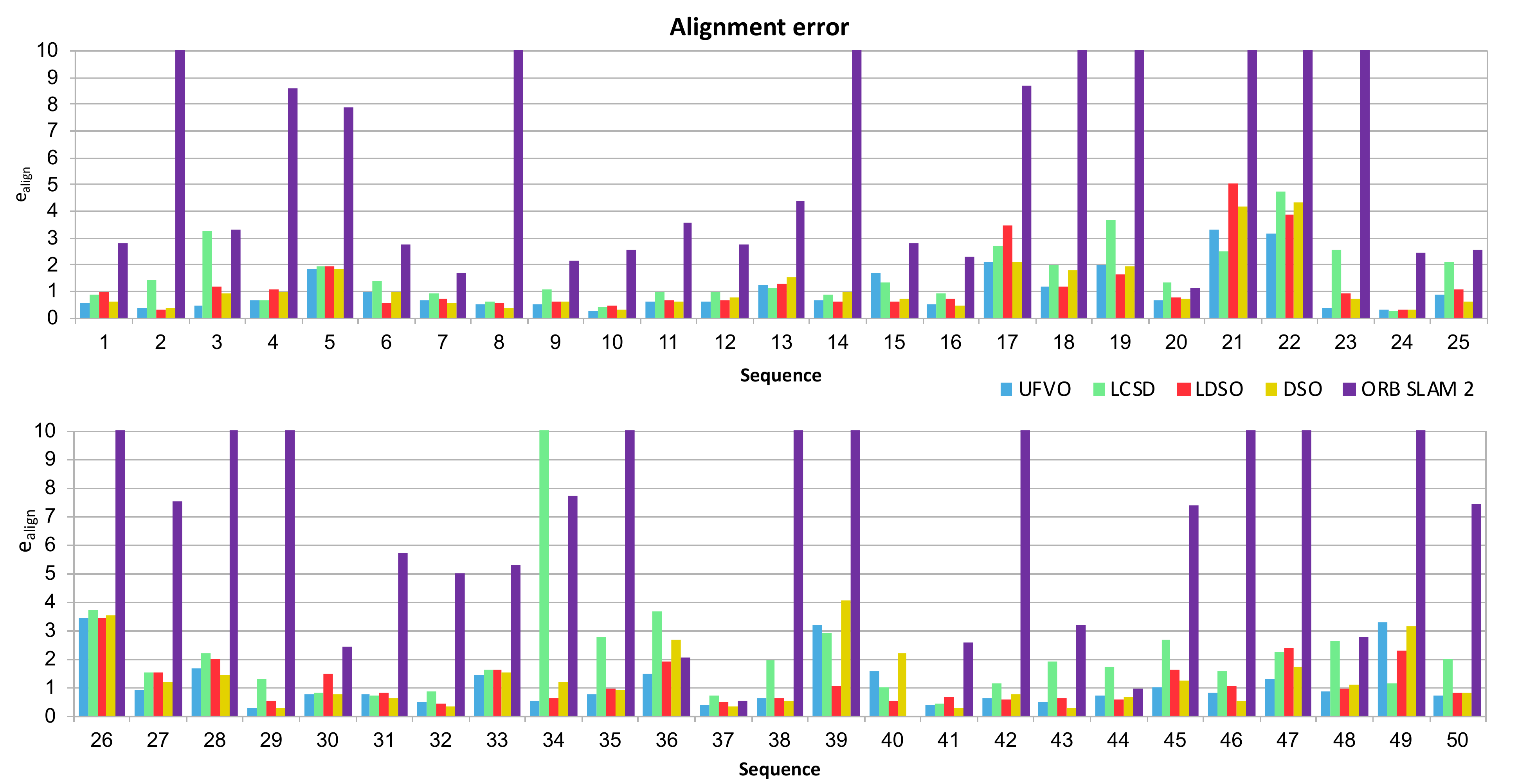}
    \caption{The alignment error $e_{align}$, as defined in \cite{engel2_2016_arxiv}, for UFVO (ours), LCSD, LDSO, DSO, and ORB SLAM 2 on all sequences from the TUM Mono dataset\cite{engel2_2016_arxiv}. The results for LDSO are taken from the extra material provided in \cite{lee_2019_ral}, DSO and ORB SLAM 2 results are taken from the extra material provided in \cite{engel_2016_ARXIV}. Finally the results of UFVO and LDSO are obtained the same way the other systems results were generated, by repeating each sequence ten times and reporting on the median. For the sake of visualization, error values beyond 10 were truncated and are not representative of the actual error.}
    \label{fig:quantitative}
\end{figure*}

\begin{figure}[!ht]
    \includegraphics[width=0.49\textwidth]{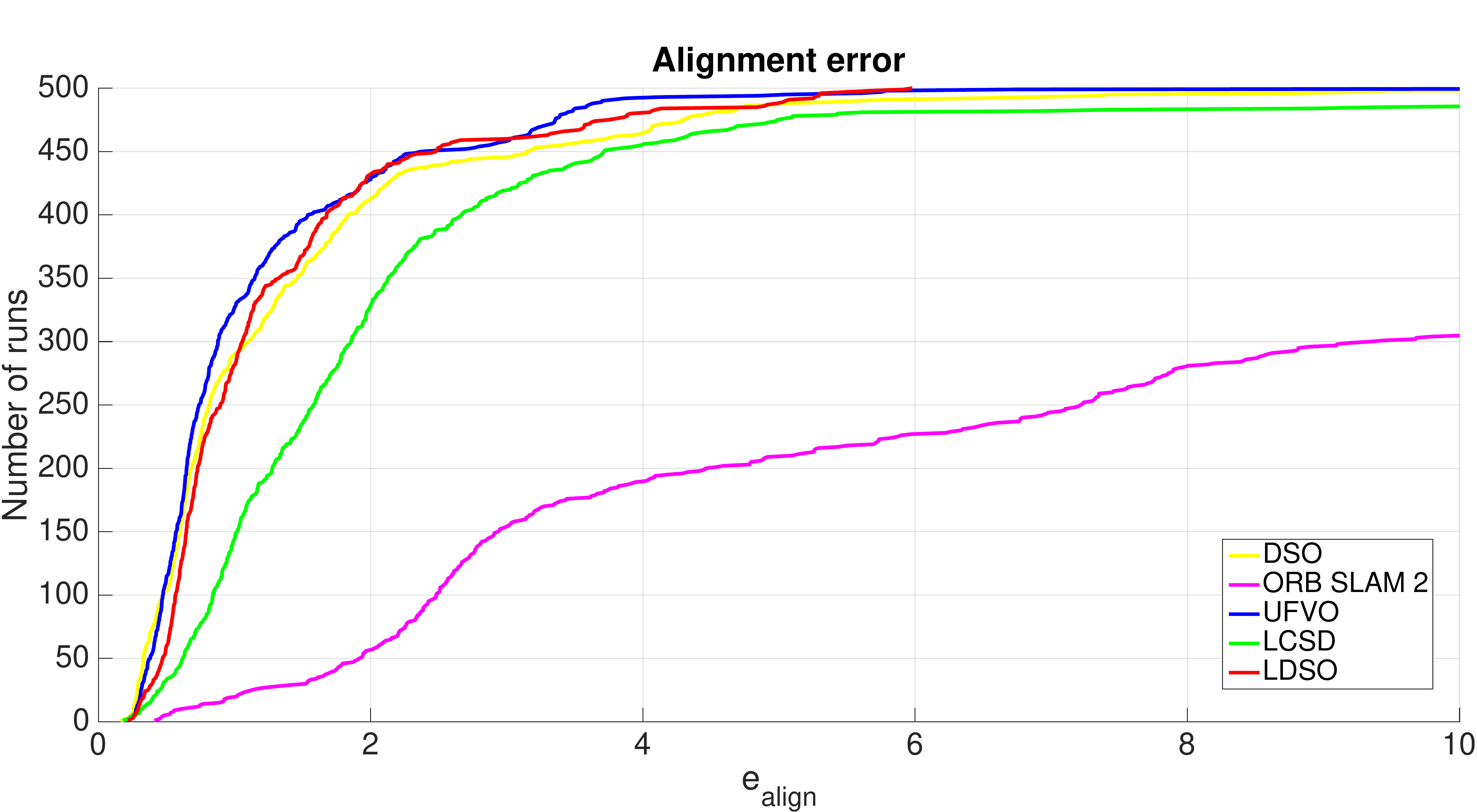}
    \caption{The number of runs for which the alignment error was lower than the alignment error shown along the x-axis. The steeper and closer to the point (0,500) the better.}
    \label{fig:alignperseq}
\end{figure}

Fig. \ref{fig:quantitative} shows the median of the alignment error over the ten runs for each sequence (the lower the better). whereas Fig. \ref{fig:alignperseq} shows the number of runs for which the alignment error (y-axis) was lower than a specific value (x-axis), the steeper the better.

\section{Discussion}
The computational cost reported in Tab. \ref{tab:CompTime} demonstrates UFVO's real-time capabilities; while tracking is relatively slower than DSO, we argue that $14$ ms per frame is well below real-time requirements, and is completely justified by the improved robustness. Furthermore, the joint optimization itself can be turned off at frame-rate, and only invoked when tracking failure is detected, thereby reducing the tracking cost to roughly the same as DSO while operating in a state similar to \cite{younes_2018_arxiv}. As for the mapping, at a mere increase of $18$ ms, we are able to generate and maintain two types of data representations which can be further exploited in a SLAM formulation. In contrast, maintaining an Indirect map alone requires at least $100$ ms per keyframe in ORB SLAM 2. 

The results reported on the alignment error (Fig. \ref{fig:quantitative} and Fig. \ref{fig:alignperseq}) reflect the complementary nature of the Direct and Indirect formulations; while the state of the art in Indirect methods (ORB SLAM2) performs consistently worse than DSO, our unified formulation was able to leverage the advantages of both paradigms to achieve a consistent improvement over both DSO and ORB SLAM 2 throughout the various sequences. 

Meanwhile, the performance of LCSD was limited in most cases to that of DSO, performing similarly or worse on most sequences. The reason for the similar results is that LCSD performs odometry using a direct formulation \cite{engel_2016_ARXIV} without any use of geometric residuals at frame-rate. However, the reduced performance compared to DSO can be attributed to three reasons: 1) first, LCSD uses DSO-reduced \cite{engel_2016_ARXIV} instead of regular DSO, which compromises accuracy for computational cost; 2) second, indirect corners suffer from reduced precision due to the discretization of the image domain to corner locations, therefore performing a geometric based optimization using Indirect corners only causes an offset around the previously found DSO optimal state; 3) third, a large number of outlier Indirect feature matches in as little as one keyframe may result in significant discrepancy between the two separate Direct and Indirect maps. All of these reasons highlight the importance of the joint formulation UFVO employs.

As for LDSO, since its ability to perform loop closure is disabled, it theoretically behaves the same as DSO, with the difference that it uses a mixture of both corners and pixels as Direct features. The immunity of corners to drift along edges, introduces performance improvement in LDSO compared to DSO; however, it does also cause worse results on some sequences, which can be attributed to the lack of a sampling mechanism that enforces homogeneous feature extraction of both types across the image like the 2D occupany grid UFVO employs. 
Furthermore, the contrast between the results of LDSO and UFVO on most sequences shows that the increased performance UFVO achieves is not due to the mere use of corner features and that UFVO is capable of using the geometric residuals to mitigate the shortcomings of its photometric counterpart.

Despite the improved performance, UFVO under-performs on a few sequences; upon closer inspection of said sequences, they all shared a significant portion of footage on highly repetitive textures (\eg , flooring carpets, grass, etc.), resulting in a large number of outlier indirect feature matches, in turn causing the reduced accuracy. While this is an inherent problem of all Indirect formulations, it was introduced to UFVO by our adaptive corner extraction mechanism that allows weak indirect corners to be extracted at such locations. In fact, increasing the lower bound on the minimum shi-tomasi score for a corner feature, caused an increase in performance by upwards of 50\% in the reported error on most of these sequences. However, doing so causes corner deprivation in weakly textured locations across other sequences. Therefore, to maintain the results coherency we evaluate all sequences using the same configuration parameters.


\section{Conclusion}
We have presented a unified formulation for Direct and Indirect features in visual odometry. By allowing corner features to have both photometric and geometric residuals, and adopting an inverse depth parametrization, we are capable of generating a single unified map in a single thread that is naturally resilient to texture-deprived environments, at a relatively small computational cost. 
Furthermore, our suggested joint pyramidal image alignment is capable of exploiting the best traits of both Direct and Indirect paradigms, allowing our system to cope with initializations far from the optimum pose, while gaining the sub-pixel accuracy of Direct methods. Finally, our point sampling and activation process ensures a homogeneously distributed, rich scene representation. While UFVO already outperforms state of the art systems, it does not make use of a global map representation, which is obtainable by maintaining a co-visibility graph over our local Indirect keyframes. Furthermore, the presence of an Indirect observation model allows for the integration of an online photometric calibration which would improve performance on non-photometrically calibrated datasets. Both topics remain part of our future work, and they are the reason why we only benchmark UFVO on the photometrically calibrated, pure path exploratory dataset (the TUM Mono dataset).



\addtolength{\textheight}{-12cm}   



\section*{APPENDIX}
\label{sec:appendix}
The photometric Jacobian $J_p\rvert_{1\times8}$ required to solve \eqref{eq:stepupdate} per pixel is found using:
\begin{multline}
J_p=\left[ \frac{f_u\nabla I_u}{d}, \frac{f_v\nabla I_v}{d}, -\frac{1}{d}(\nabla I_uuf_u+\nabla I_vvf_v),\right.\\  -\nabla I_vf_v(1+v^2)-\nabla I_uf_uuv, \nabla I_uf_u(1+u^2)+\nabla I_v f_v uv, \\\left. \nabla I_vf_vu-\nabla I_uf_uv,-\frac{t_je^{a_j}}{t_ie^{a_i}}(I_i[\mathbf{p}]-b_i),-1  \right]
\end{multline}
where $u$ and $v$ are the coordinates of the point $\mathbf{p'}$ in the new frame's image plane, $f_u$ and $f_v$ are the focal length parameters, and $\nabla I$ is the directional image gradient.
The geometric Jacobian $\mathbf{J}_g\rVert_{2\times8}$ per corner is computed as:
\begin{equation}
\mathbf{J}_g =
\begingroup 
\setlength\arraycolsep{2pt}
\begin{bmatrix}
\frac{f_u}{d},&0, &-\frac{f_uu}{d}, &-f_uuv, &f_u(1+u^2),&-f_uv,&0,&0 \\
0,&\frac{f_v}{d},&-\frac{f_vv}{d},&-f_v(1+v^2),&f_vuv,&f_vu,&0,&0
\end{bmatrix} 
\endgroup 
\end{equation}


\bibliographystyle{IEEEtran}
\bibliography{younes_bib}

\end{document}